\pdfoutput=1
\documentclass[11pt,hyphens]{article}

\usepackage{ACL2023} %add[review] before {ACL2023} for review version

\usepackage{times}
\usepackage{latexsym}
\usepackage{xcolor}
\usepackage[T1]{fontenc}
\usepackage[utf8]{inputenc}

\usepackage{microtype}

\usepackage{inconsolata}
\usepackage{hyperref}
\usepackage{amsmath}

\title{Position Paper\\
\vspace{0.5cm}
Losing our Tail, Again: (Un)Natural Selection \& Multilingual LLMs}

\author{Eva Vanmassenhove \\
  Research Centre for Cognitive Science \& Artificial Intelligence \\
  Department of Computational Cognitive Science\\
  Tilburg University, The Netherlands \\
  \texttt{e.o.j.vanmassenhove@tilburguniversity.edu} \\
}  %\And
\begin{document}
\maketitle
\begin{abstract}
\noindent
Multilingual Large Language Models considerably changed how technologies influence language. While previous technologies could mediate or assist humans, there is now a tendency to \textit{offload} the task of writing itself to these technologies, enabling models to change our languages more directly. While they provide us quick access to information and impressively fluent output, beneath their (apparent) sophistication lies a subtle, insidious threat: the gradual decline and loss of linguistic diversity. In this position paper, I explore how model collapse, with a particular focus on translation technology, can lead to the loss of linguistic forms, grammatical features, and cultural nuance. Model collapse refers to the consequences of self-consuming training loops, where automatically generated data (re-)enters the training data, leading to a gradual distortion of the data distribution and the underrepresentation of low-probability linguistic phenomena. Drawing on recent work in Computer Vision, Natural Language Processing and Machine Translation, I argue that the many \textit{tails} of our linguistic distributions might be vanishing, and with them, the narratives and identities they carry. This paper is a call to resist linguistic flattening and to reimagine Natural Language Processing as a field that encourages, values and protects expressive multilingual diversity and creativity.
\end{abstract}

\section{A Simple Evolutionary Trade-Off}

A few million years ago, we lost our tails. Once useful (and presumably quite fun), they became obsolete, absorbed by evolutionary trade-offs. They disappeared through natural selection. Darwin already drew parallels between the evolution of languages and those of species noting that they follow `curiously parallel' paths in The Descent of Man \citep{darwin1871descent}. He went so far as to suggest that ``\textit{the survival of certain favored words in the struggle for existence is \textbf{natural selection}}'' \citep[p.~61]{darwin1871descent}, highlighting how language itself is subject to various evolutionary forces. However, it is unlikely that even Darwin could have anticipated the current path of language evolution and the latest forces that have come into play. 

Today, it is not just `natural'\footnote{While the technological forces that shape language can certainly be seen as part of the `natural' forces influencing our linguistic ecosystem, there is a crucial shift when we move from tools that help us record, polish, or disseminate language to models that actively \textit{generate} it. In such systems, the technology becomes not just a medium but a core driver of language change. I will return to this later in the paper.} forces but technological ones that exert strong selective pressures on languages and influence the struggle for linguistic survival on various levels. Multilingual Large Language Models (LLMs) are trained as next-word predictors over large datasets. As a result, they tend to amplify statistically likely languages and linguistic forms, whether these are words, subwords, or syntactic constructions, thereby potentially pruning away our rich, long statistical tails. \\
Will a technology-driven, artificial selection merely accelerate an already inevitable trajectory, or does it mark a more disruptive turn? In other words, does losing our `language' tails imply the disappearance of obsolete, fun, and decorative elements or does it signal a more fundamental erosion of diversity and richness? In this paper, I take a position on these questions, arguing that the LLM-era marks a disruptive turn: unlike `natural' evolution, technological selection in current mainstream LLMs (under prevailing data and training regimes) tends to reduce rather than reshape linguistic diversity.

\paragraph{A quick note on `language tails':}
In language, as in other complex systems, many distributions are long-tailed: a few elements occur very frequently, while most are rare. These ``tails'' appear across multiple linguistic levels. For example, a word like `the' appears more often than `disperse' or `eloquent'. Yet, these low-frequency terms are grammatical, meaningful and important within specific domains or contexts. Similar long-tailed patterns can be observed across morphology, syntax, phonology, as well as at the cross-linguistic level. 

% \paragraph{A quick note on language tails} Like many complex systems, language follows long-tailed distributions: a few elements occur very frequently, while most are rare. 

% A word like `the' appears far more often than `disperse' or `eloquent', yet all are grammatical and meaningful within their contexts. Such tails occur across linguistic levels (morphological, syntactic, cross-linguistic).

\section{\textit{Natural} Language Selection}\label{sec:natural}

Before turning to how technology reshapes language, it is helpful to briefly consider how languages evolve `naturally'. Evidently, this is not meant to be a comprehensive account of language evolution since doing justice to that topic requires several volumes.\footnote{For readers interested in a broader, more comprehensive and nuanced history on, among others, the evolution of language, I recommend ``The Language Puzzle''~\cite{mithen2024language} and ``How Language Began''~\cite{everett2017language}.} The goal here is to highlight a few key aspects of how humans learn and transmit language, to lay the groundwork for drawing parallels with how Multilingual LLMs learn, transmit, and affect language. Broadly speaking, natural selection in languages operates at two main, related levels: \textit{across languages} (Section~\ref{subsec:across}), where some languages survive and thrive while others gradually disappear, and \textit{within languages} (Section~\ref{subsec:within}), where certain words, expressions, or grammatical structures persist over time while others fall out of use. 

\subsection{Across Languages}\label{subsec:across}
Historically, population shifts, cultural dominance, and environmental factors have repeatedly driven language extinction~\cite{bromham2022global}. Nearly half of the world's roughly 6,000 languages\footnote{Estimates vary between 3,000 and 10,000~\cite{crystal2002language}.} are endangered, many disappearing at an alarming rate.\footnote{\url{https://www.unesco.org/en/articles/multilingual-education-bet-preserve-indigenous-languages-and-justice}.} While language loss is often viewed as something tragic, it can stem from a shared communicative goal. For instance, speakers might adopt a dominant language to improve mutual understanding. Evidently, even in these seemingly positive scenarios, something linguistically and culturally valuable is lost.

More subtly and at the same time more \textit{universally}, we might also be witnessing a different type of loss: the loss of richness and diversity \textbf{within} languages themselves. Such a loss is less likely to be driven by collective needs like mutual intelligibility\footnote{Of course, in some  specific cases or contexts, simplification can help communication across a more diverse set of speakers.}, but can most likely be attributed to converging pressures and constraints from language technologies. Therefore, I want to turn the attention more inwards, and focus on the erosion of diversity \textit{within} languages.

\subsection{Within Languages}\label{subsec:within}

Whereas the extinction of languages is often driven by external forces, changes within a language are also shaped by internal cognitive bottlenecks. The acquisition of a language, for instance, happens under rather constrained conditions: humans are exposed to sparse, ambiguous, and often noisy and imperfect input—sometimes discussed under the notion of a “poverty of stimulus” \cite{chomsky1980rules}. Additionally, we face cognitive limitations including finite memory, a limited attention span, and bounded processing capacities. Yet, despite these many hurdles (or perhaps because of them, as suggested by \citet{decaro2008individual}) dominant languages continue to thrive.

% Language can be seen as a cultural technology \cite{Gopnik2017}, a tool that helps us pass on knowledge from generation to generation. We acquire and leverage this tool through these rather constrained conditions and are exposed to sparse, ambiguous and often imperfect input. In addition to this so-called \textit{poverty of stimulus} \cite{chomsky1980rules}, we face cognitive limitations such as finite memory, a limited attention span and bounded processing capacities. Yet, despite these many hurdles (or perhaps because of them\cite{decaro2008individual}) our language continues to thrive.

In the following sections, I will distinguish between two types of language-internal changes: (i) the emergence of structure, briefly revisiting some foundational research on grammar and compositionality (Section~\ref{subsec:structure}); and (ii) the evolution of our lexicon, that is, how words are lost, gained, or transformed over time (Section~\ref{subsec:lexicon}). While in reality, these processes are to some extent intertwined, they are shaped by different pressures.

% In what follows, I will distinguish between two types of language-internal change: the evolution of structure—how compositional and grammatical systems emerge and adapt—and the evolution of the lexicon—how words, subwords, and less frequent constructions are gained, lost, or transformed. While these processes are intertwined in practice, they are shaped by different pressures and operate on different levels: structural change tends to reflect pressures toward expressivity, learnability, and compositionality, whereas lexical change is more directly exposed to usage frequency, cultural shifts, and increasingly, technological filtering and amplification.

% For humans, inductive biases like a preference for compressibility, learnability, simplicity, or efficiency Kirby et al. (2015); Tamariz and Kirby (2015); Gibson et al. (2019) naturally shape how languages evolve. Some even argue that humans’ cognitive limitations may be beneficial for language acquisition DeCaro et al. (2008); Poletiek et al. (2018).

\subsubsection{Towards Structure}\label{subsec:structure}
% To study how specific properties of language emerge through cultural evolution, a popular framework is \textit{iterated learning}~\cite{brighton2005language}. It is a hypothetical procedure that studies behaviour in a chain of intelligent agents that acquire knowledge (imitation) and pass it onto to a next generation (transmission)~\cite{ren2024bias}. In short, iterated learning refers to the process where an individual’s knowledge or skill is developed by observing the behavior of another person, whose behavior itself reflects their own knowledge or competence \cite{smith2003iterated}.

%shortened version
Iterated learning~\cite{brighton2005language,smith2003iterated,ren2024bias} models how linguistic structure emerges through repeated cultural transmission, where each generation of learners infers patterns from the behavior of the previous one. 

To cope with limited and noisy input, humans rely on (implicit) inductive biases such as simplicity and compressibility to learn a language~\cite{kirby2015compression}. These structured preferences guide learning by enabling pattern extraction and structural generalization from sparse data. Research by~\citet{smith2003complex} showed that when language is transmitted under cognitive and communicative pressure, structure and compositionality emerge through repeated cultural transmission cycles -- processes shaped by our cognitive biases that favor more learnable and generalizable patterns. Far from being detrimental, such biases are often considered essential~\cite{decaro2008individual}, allowing generalization without linguistic impoverishment.
% or collapse.

%NOT SUMMARIZED VERSION
% To cope with the limited and noisy input, humans rely on (implicit) inductive biases like simplicity and compressibility in order to learn language ~\cite{kirby2015compression}. These structured preferences or heuristics help guide us through our learning process relying on limited and noisy training data. They enable us to extract patterns and find structure. Research by ~\citet{smith2003complex}, for instance, illustrated that when language is transmitted under cognitive and communicative pressure, structure and compositionality emerge. Syntax, for example, can be viewed as arising through repeated cycles of cultural transmission. A process shaped by our cognitive biases that favor more easily learnable and generalizable patterns. These inductive biases are not detrimental to language learning. On the contrary, they are sometimes said to be essential~\cite{decaro2008individual} as they allow us to generalize without causing a language impoverishment or a collapse. 

This also implies that our learning process involves \textit{convergence}. Convergence towards systematic forms, towards structure and patterns. This convergence is not of the kind that leads to a loss of expressivity. It is a structural, higher-level convergence that supports more open-ended productivity, one where nonsense syllables can become, for instance, compositional morphemes. 

While iterated learning and cultural transmission help explain how structure in language becomes more regular and learnable over time, not all aspects of language evolve in such a way. Our lexicon, the system of words and meanings, often changes in more irregular and dynamic ways. Vocabulary is more susceptible to social influence, borrowing, innovation, and drift. 
\subsubsection{Lexical Changes}\label{subsec:lexicon}
%stochastic drift -> genetic drift)

Despite our limited memory, languages are and remain remarkably rich. Words may disappear, fall out of fashion, or decrease in usage, yet their overall vocabularies remain relatively stable, often even expanding when languages flourish driven by social needs, interactions, scientific progress, or creativity. For instance, based on over five million digitized books,\footnote{The 2010 analysis covered roughly 4\% of all books ever published.} \citet{michel2011quantitative} estimated that English grew by more than 70\% over the past fifty years, adding on average about 8500 new words per year. A later study by \citet{gerlach2013stochastic} showed that vocabulary growth largely scales with corpus size, with the rate of new-word introduction slowing but never vanishing as datasets expand. These findings can, however, be contrasted with claims of convergence towards a maximum vocabulary size~\cite{bernhardsson2009meta}.

% the number of words in English based on a corpus of 5,195,769 digitized books.\footnote{Their quantitative analysis dates from 2010 and back then accounted for approximately 4\% of all books ever published.} They observed an increase in vocabulary over the past century with an addition of approximately 8500 words per year, increasing the vocabulary size by over 70\% the past 50 years. However, a more recent study on the same database by \citet{gerlach2013stochastic} found that vocabulary size is mainly driven by the size of the available corpus, with the rate of introduction of new words decaying but never vanishing as corpus size grows. This can contrasted with other research that claims there is a convergence towards a maximum vocabulary size \cite{bernhardsson2009meta}.

The vocabulary growth reported for English in \citet{michel2011quantitative} and \citet{gerlach2013stochastic} aligns with findings showing that languages with larger speaker populations tend to have larger vocabularies~\citep{reali2018simpler}. At the same time, widely spoken languages often exhibit simpler morphology (fewer inflections) while smaller communities may develop structurally more complex systems~\citep{lupyan2010language}.

% The vocabulary growth reported for English by \citet{michel2011quantitative} and \citet{gerlach2013stochastic} is in line with findings that languages with larger speaker populations tend to have larger vocabularies as opposed to smaller ones \citep{reali2018simpler}. At the same time, languages spoken by many people tend to be structurally simpler, exhibiting fewer cases and less inflections (e.g. English); while smaller communities sometimes develop languages with more structural complexity \citep{lupyan2010language}. 

Word survival is frequently influenced by positive frequency-dependent selection~\cite{pagel2019dominant}, yet frequency alone does not determine a word’s fate~\cite{altmann2011niche}. Competition among words is also shaped by cognitive and communicative pressures that favor efficiency and clear categorization. Naming systems for colours or emotions, for example, tend to converge across cultures, reflecting shared communicative needs rather than pure frequency effects~\cite{petersen2012statistical}. Even common words can disappear if they no longer align with prevailing cognitive, social, or technological forces. Consider the case of `radiogram', `Roentgenogram', and `X-ray', once competing terms for the same concept~\cite{petersen2012statistical}. Although `Roentgenogram' dominated early scientific discourse, it was gradually replaced by the shorter and more efficient `X-ray'~\cite{piantadosi2011word}, a shift further reinforced by the global rise of English as the scientific lingua franca.

% What is interesting though, is that the core vocabularies of languages, typically defined as a set of basic, essential words, remain relatively stable over time within languages. These vocabularies are furthermore cross-linguistically speaking similar in size, shaped by some universal human experiences although their exact composition varies with cultural and environmental context. 

Vocabulary, in short, is shaped not just by usage frequency but by communicative efficiency, cognitive constraints, technological mediation, and sociocultural forces, reflecting a dynamic and context-sensitive interplay between convergence and divergence. Yet, as technological mediation increasingly shapes communication, these dynamics may come to be governed by a different kind of selection.

\section{A Force of \textit{Technology}}\label{sec:technology}

Darwin argued that languages evolve through a process of variation and selection, convergence and divergence, much like species. We have seen how such dynamics give rise to compositionality and structure, as well as changes in our lexicon. Broadly speaking, we have seen that language reflects a tension between convergence and divergence. However, we set out to explore what happens when the strongest selective forces are no longer human cognition or social need, but instead arise from (large-scale) technological forces. 

First, we briefly examine the relationship between language and cultural technologies that predate the rise of (L)LMs (Section~\ref{subsec:culturaltech}). This provides context for the \textit{Descent of Language} section (Section~\ref{subsec:descent}), where I examine early signs of linguistic decline in computational models and discuss more recent work on LLMs.

\subsection{Language \& Cultural Technologies}\label{subsec:culturaltech}
It is important to note that it is not the first time that technology has played a disruptive role in shaping language. Language has never existed in isolation and (cultural) technologies have continuously influenced how language is transmitted. While it may be hard to imagine today, even \textit{writing} was once considered a disruptive technological innovation:
\vspace{5pt}
\begin{quote}
    \textit{They seem to talk to you as though they were intelligent, but if you ask them anything about what they say from a desire to be instructed they go on telling just the same thing forever.}\\
    \vspace{-4pt}
    \hfill --- Plato, \textit{Phaedrus} 275d\nocite{Plato2002}
\end{quote}
% \vspace{-2pt}
As cognitive scientist Alison Gopnik pointed out in her 2023 ACL keynote \citep{gopnik2023acl}, without any context, this critique of \textit{writing} could just as well apply to LLMs today. There is a surface-level fluency that (over)confidently mimics understanding, yet lacks genuine responsiveness, real-world grounding, or explanatory depth. While we no longer worry about the \textit{dangers} of writing, history reminds us that each new `tool' reshapes how language is used, transmitted, and transformed. Some languages, the so-called "long-tailed" ones,\footnote{Languages that receive limited localization attention or commercial investment; this does not always correspond to speaker numbers. For instance, Bengali is the $7^{\text{th}}$ most spoken language, but falls outside the top 50 most localized languages \citep{lionbridge2020longtail}.} have been particularly impacted. Their marginalization has been reinforced by successive cultural technologies (from writing to the internet), and by their exclusion from the digital sphere. This phenomenon, described as \textit{digital language death}, is said to affect roughly 95\% of the world’s languages~\cite{kornai2013digital}.

% might not be concerned anymore about the potential side-effects or dangers of \textit{writing} as a tool, it is worth reflecting on the fact that, throughout history, various technologies have influenced how we use, transmit, and transform language to different extents. 

% Looking across languages, the loss of languages with few speakers or limited commercial value (often referred to as `long-tailed languages'\footnote{Long-tailed languages refer to languages that receive little localization attention or commercial investment. That does not necessarily imply that they have a small number of speakers. For instance, Bengali is the $7^{\text{th}}$ most spoken language, but falls outside the top 50 most localized languages \url{https://www.lionbridge.com/blog/translation-localization/localizing-long-tail-languages/}}) has been exacerbated by the emergence of certain cultural technologies (e.g. writing, the emergence of the internet...). Beyond how technologies shape language use, the absence or the inability to `ascend' to the digital realm can be equally telling. This phenomenon has been described as \textit{digital language death} and affects 95\% of our languages~\cite{kornai2013digital}. 

Earlier technologies also marked turning points in the evolution of language. The printing press, for instance, helped standardize spelling and grammar, elevating dominant dialects while marginalizing others~\cite{sasaki2017publishing}. More recently, automatic spell-checkers boosted the `reproductive fitness' of recognized forms at the expense of other alternatives, contributing, for example, to the rising dominance of `colour' over `color'~\cite{petersen2012statistical}.\footnote{Amusingly, the tool I am writing in still nudges me away from `colour'.}

% Earlier technologies also marked key turning point in the evolution of language. The invention of the printing press, for example, contributed to the standardization of spelling and grammar, promoting dominant dialects while marginalizing others~\cite{sasaki2017publishing}. More recently, automatic spell-checkers led to a boost in `reproductive fitness' for recognized words at the expense of (unstandardized) alternatives~\cite{petersen2012statistical}, as seen in the increased popularity of `colour' over `color'~\cite{petersen2012statistical}\footnote{Interestingly, the tool I'm currently using attempts to discourage me from using `colour'.}.

While the impact of technology on language and concerns regarding its disruption are not new, the current wave of generative models operates at a different level, scale, and speed. They represent a new kind of intervention, one in which technology is not simply storing, mediating, or `correcting' language, but actively \textbf{generating} it across different domains, tasks, and platforms. In a sense, the technology quite suddenly moved from the passenger's seat to the driver's seat. 

This shift in agency may have important implications for languages and their evolution. Concerns regarding the loss of diversity \textit{across} languages due to recent technologies, such as LLMs, have been explicitly raised in recent seminal work within the field of Natural Language Processing (NLP) by \citet{joshi2020state} and \citet{bender2021dangers}, among others. Here, I shift the focus inward, to the internal diversity \textit{within languages}. This dimension has received comparatively little attention, even though current multilingual LLMs exert significant influence on what is preserved, erased, or amplified within a language itself. 

% Importantly, these potential pressures may not be limited to obvious levels such as vocabulary or stylistic choices, but could also emerge in less expected ways—at morphosyntactic levels, through non-linguistically motivated subword or character segmentations, or via subtle “leakage” of dominant language structures. For example, one might hypothesize that multilingual LLMs could begin to favor English-like constructions or idiomatic patterns even in typologically distant languages. This dimension has received comparatively little attention so far, even though such effects, if present, could significantly shape what is preserved, lost, or amplified within a language.

% \dimitar{From ChatGPT: So, yes, language technologies are cultural technologies, but how they're designed and used determines whether they enrich or erode culture.}
% \textcolor{teal}{MOVE ELSEWHERE: Language is a form of cultural technology, a \textit{tool} that enables the transmission of knowledge across generations~\cite{Gopnik2017}. Despite our constraints and the (generational) transmissions, it is worth highlighting that language does not collapse or degrade over time. This stands in contrast to what can be observed when language models are trained on their own, or other models', output.} 

\subsection{Large Language Models}
LLMs mark a significant shift in how language is produced, accessed, and reused, and perhaps more interestingly, how the cognitive labor involved in writing and idea formulation is now often \textit{offloaded} to technology. Earlier language technologies were built for (domain-) specific tasks; today's foundation models, however, operate across domains and applications. Large-scale models are rapidly becoming an integral part of a broad range of our everyday activities~\citep{bick2026rapid}, allowing them to shape and influence language more directly than before. Unlike earlier tools that assisted human writing, these models now drive change by producing language themselves.

Emerging research, however, suggests that LLMs introduce subtle, yet cumulative, distortions \cite{shumailov2023curse}. While initially \textit{imperceptible}, these small changes can accumulate across multiple training cycles, leading to a phenomenon coined \textit{model collapse}, where models trained on their own outputs progressively lose quality and diversity. In computer vision, such a collapse leads to \textit{visible} artefacts in AI-generated images \citep{alemohammad2023self}, yet for language, the consequences remain underexplored. This is despite the fact that language shapes and possibly constrains human thought, meaning that this gradual impoverishment or distortion of linguistic output could have profound and far-reaching implications.

LLMs will likely continue to substantially influence both the content we are exposed to (images, texts, audio, etc.) and the systems that generate this content given that their output will increasingly re-enter the training cycle.\footnote{A substantial portion of multilingual web content has been shown to consist of machine-translated text, indicating that language distributions are already significantly shaped by automated systems \citep{thompson-etal-2024-shocking}.} Therefore, at this point we can assume that interactions between models are not hypothetical but inevitable \citep{martinez2023towards}. Such interactions can occur through (partial) training on output from another LLM or a model's own output \citep{renbias}. The subsequent feedback loops this creates, where models learn from their own output or that of others, accelerate concerns regarding the distortion of language. 

So far, research efforts have largely focused on sustaining the benefits of training from large-scale, human-generated data scraped from the Web, summarized in this recent article as: ``\textit{LLMs' world is our word}''\footnote{\url{https://www.theguardian.com/technology/article/2024/sep/07/if-journalism-is-going-up-in-smoke-i-might-as-well-get-high-off-the-fumes-confessions-of-a-chatbot-helper?utm_source=chatgpt.com}}. But perhaps more concerningly: ``\textit{Our world} might be turning into \textit{their word}''. 

% In the next section, we briefly review recent empirical evidence that suggests these dynamics are already contributing to linguistic convergence and loss of diversity.

% Since MLLMs will likely continue to substantially influence both the content we consume (images, texts, audio) and the systems that generate this content, interactions between models are not hypothetical but inevitable. These interactions occur through (partial) training on output from another MLLM or even on a model's own output \citep{renbias}. Such feedback loops where models learn from their own output or that of other models, accelerate concerns regarding \textit{model collapse} and the distortion of language. So far, emphasis was put on sustaining the benefits of training from large-scale data scraped from the web and the value of human-generated data for training, aptly summarized in this recent article as: \textit{LLMs world is our word}\footnote{\url{https://www.theguardian.com/technology/article/2024/sep/07/if-journalism-is-going-up-in-smoke-i-might-as-well-get-high-off-the-fumes-confessions-of-a-chatbot-helper?utm_source=chatgpt.com}}. More concerningly, however, the reverse may also be true: \textit{Our world} is increasingly turning into \textit{their word}. 

\subsection{The \textit{Descent} of Language}\label{subsec:descent}

% Recent studies expose a concerning pattern, one of \textit{convergence}, and potential collapse. Terms like model collapse~\citep{shumailov2023curse}, Model Autophagy Disorder (MAD)~\cite{alemohammad2023self}, and Habsburg AI'\footnote{A term that was coined by Jathan Sadowski.} have recently been coined to refer to the degeneration of generative models as they are increasingly trained on their own outputs. Each generation of the model becomes less expressive and structurally weaker, like a form of linguistic inbreeding that might eventually lead to a language equivalent artefact of the Habsburg jaws. Already before these terms gained traction, earlier research empirically showed how neural models indeed amplify dominant linguistic forms while forgetting or flattening rarer, low-probability ones.

Even before terms such as `model collapse'~\citep{shumailov2023curse}\footnote{As recently discussed in \citet{schaeffer2025position}, model collapse encompasses a broad range of phenomena; here, we explicitly focus on the distributional aspect, which relates to the erosion of long-tail linguistic diversity central to our argument.}, `Model Autophagy Disorder' (MAD)~\cite{alemohammad2023self} or `Habsburg AI'\footnote{A term coined by Jathan Sadowski (\url{https://x.com/jathansadowski/status/1625245803211272194?lang=en}).} gained traction, earlier research empirically showed how statistical language models\footnote{Whether they are called statistical, neural or large language models, in the end, they are still all statistical models.} indeed amplify dominant linguistic forms while forgetting or flattening rarer, low-probability ones.

% Terms like model collapse~\citep{shumailov2023curse}, Model Autophagy Disorder (MAD)~\cite{alemohammad2023self}, and Habsburg AI'\footnote{A term that was coined by Jathan Sadowski.} have recently entered the discourse to describe a similar underlying phenomenon: the degeneration of generative models as they are increasingly trained on their own outputs. Like a form of linguistic inbreeding, each generation of the model becomes more uniform, less expressive, and structurally weaker—hence the nod to the infamous Habsburg jaw. Before these metaphors took hold, however, earlier research had already shown how neural models amplify dominant linguistic forms while flattening or forgetting rarer, low-probability ones.

% the degeneration of generative models as they are increasingly trained on their own outputs. Like a linguistic version of inbreeding, each generation of the model becomes more distorted, more uniform, and less robust

\subsubsection{Precursor: Statistical and Neural Translation Models}
As a precursor to this line of work, studies in Machine Translation (MT) \citep{vanmassenhove-etal-2019-lost,vanmassenhove2021machine} provided empirical evidence that both statistical and neural models systematically favor frequent lexical and morphological patterns, reducing linguistic diversity. This raises two concerns: (i) technically, frequency bias diminishes lexical richness and can eliminate infrequent but grammatically necessary forms; and (ii) sociolinguistically, machine‐generated translationese may, over time, influence language itself~\citep{kranich2014translations}. For instance, \citet{vanmassenhove2021machine} showed that MT systems disproportionately produce masculine \textit{pr\'esident} over feminine \textit{pr\'esidente} when translating English \textit{president} into French, reflecting data imbalances. Under iterative training, rare forms may disappear entirely: in their experiments, the plural \textit{pr\'esidentes} vanished. Such low‐probability forms are not noise but carriers of grammatical precision, expressiveness, and social meaning. Yet, current models fail to distinguish these cases, discarding rare forms indiscriminately. 

While the previous examples concerned lexical and morphological variation, similar effects arise at the morphosyntactic level. \citet{luo2024diverge} found MT outputs to be structurally closer to the source text than human translations, with fewer morphosyntactic divergences. Beam search biases toward “safe,” high‐probability constructions, reinforcing convergence and reducing syntactic diversity.

% \citet{luo2024diverge} reached a similar conclusions but focused on more fine-grained morphosyntactic patterns when comparing MT to human translations (HT). They observed a significant difference between HT and MT with MT producing more one-to-one alignments. In other words, the automatically generated translations are structurally `closer' to the source when compared to the HT ones. They attribute this convergence to the beam search, which biases MT towards less diversity in the outputs. Aside from that, they also show that many morphosyntactic divergent patterns in HT are correlated with an overall decrease in MT performance. 

% observing convergence and less diversity in the MT generated translations.

% While our study reaches a similar conclusion that MT is less diverse than HT, we explore morphosyntactic patterns on a more fine-grained level, and also reveal the bias of MT (and more specifically beam search) towards convergent structures.

% This idea is somewhat comparable to the more well-known issue of \textit{sealing} knowledge \citep{lindemann2023sealed}, happening to non-dominant knowledge in search engines, but then on a more micro-level in terms of language itself. 
\vspace{-0.2cm}
\subsubsection{Generative Models}
In recent work on LLMs, concerns about \textbf{model collapse} have gained traction: when models are trained on data increasingly composed of their own (or other models') output, their behavior begins to \textit{converge}, potentially degrading over time~\cite{shumailov2023curse}. In computer vision, this degeneration has been made \textit{visible}, with iterative training cycles producing recognizably distorted `Habsburg Jaw' artifacts~\cite{alemohammad2023self}. In language, such collapse is likely subtler and harder to detect, yet potentially more consequential, given the foundational role of language in human cognition and cultural evolution.

Empirical studies suggest early signs of such dynamics. \citet{mccoy2023embers} show that even for simple deterministic tasks, LLM behavior is sensitive to probability: GPT-4’s performance drops dramatically when the correct output sequence is low-probability, revealing \textit{embers of autoregression}.
% They argue for evaluations grounded in model training objectives and constraints: a \textit{teleological} perspective.
Evidence from real-world text production points in a similar direction, reflecting probability-driven effects. \citet{kobak2024delving} report sharp spikes in terms such as \textit{delve}, \textit{crucial}, and \textit{significant} in PubMed abstracts following the release of public LLM tools, exceeding even pandemic-related shifts (e.g. the sudden spike of \textit{corona}). This suggests subtle, distributional nudging of academic discourse toward statistically likely phrasing.

At the distributional level, \citet{shumailov2023curse} demonstrate how low-probability events disappear first in iterative training, with models gradually converging toward high-probability sequences. Extending these observations to large-scale generative settings, \citet{jiang2025artificial} identify an ‘Artificial Hivemind’ effect, characterized not only by repetition within a single model, but also by striking homogeneity across different models when responding to diverse open-ended prompts. 

Importantly, these dynamics are not limited to earlier generations of LLMs and may even be amplified by Reinforcement Learning from Human Feedback \citep{kirk2023understanding} or Supervised-Fine-Tuning, where aggressively updating the model towards observed data distributions can further entrench the underrepresentation of low-probability forms. Recent work on instruction-tuned LLMs further shows that such constraints can manifest in practice as what is referred to as `diversity collapse', with models converging toward semantically similar responses \citep{yun2025price}.\footnote{These trade-offs are often discussed under the notion of the `alignment tax' \cite{bai2022constitutional,ouyang2022training}.}

Focusing more on linguistic diversity, \citet{guo2024benchmarking} leverage a comprehensive evaluation framework to assess LLM outputs and observe a significant decrease in diversity when comparing human language to LLM-generated content, especially when focusing on tasks that require creativity. They depict a rather complex pattern where instruction tuning does improve lexical diversity, but this comes at a cost, as it narrows the expressive flexibility seen by a decrease in overall syntactic and semantic diversity.

% Similarly, \citet{shumailov2023curse} who coined the term \textit{model collapse}, illustrates how models converge over generations. Their experiments indicate that at first, the tails of our data (the low-probability events) begin to disappear. Over time they observed a convergence toward more probable token sequences. The authors link these observations to earlier `poisoning' phenomena seen in web search driven by humans who attempted to misguide social networks or search algorithms. For search engines, this led to interventions based on filtering out or downgrading farmed articles\footnote{\url{https://googleblog.blogspot.com/2011/02/finding-more-high-quality-sites-in.html}}. However, with LLMs it is less clear how we can filter out such content to prevent models from collapsing over time \cite{shumailov2023curse}. 

% Focusing more on the linguistic aspects of language, \citet{guo2024benchmarking} leveraged a comprehensive evaluation framework to assess the linguistic diversity of LLMs and observed a significant decrease of diversity when comparing human language to LLM-generated content, especially when focusing on tasks that require creativity. They depict a rather complex pattern where instruction tuning does improve lexical diversity but it comes at a cost since it narrows the expressive flexibility seen by a decrease in overall syntactic and semantic diversity.

\section{Thoughts \& Discussions}
Darwin noted the similarities between the evolution of species and that of languages. Of course, he was referring to mechanisms of gradual change through natural evolution. I could, however, not help be amused by the rather unexpected parallel that can be drawn between the evolution of our species and the effect of recent technologies on languages. It seems that, once again, our species and language are following a similar path: \textit{We are losing our tails}. While our physical appendages became obsolete and even cumbersome, I argue that its language equivalent is anything but that, capturing the rich diversity and continuous evolution of language, shaped both by the need to articulate novel concepts and by our ongoing desire to signal belonging and distinction within social groups.

While one could argue that across languages, various natural forces, sometimes in combination with technological ones, have contributed to a significant loss of diversity across languages, leading even to \textit{language death}, when we shift the focus to the internal diversity within languages, natural evolution has led to structure, compositionality, and a growth in terms of vocabulary, at least for thriving languages. This stands in contrast with what we observe when language is \textit{driven} (largely or solely) by technological forces, whose internal pressures seem to lead to a reduction in expressivity, creativity, and a loss of overall lexical diversity regardless of whether the language is flourishing or not. 

In the paragraphs below, I set out my reflections and formulate a position on what it means when current technological forces become the main engines of language change.

\paragraph{(L)LMs reduce linguistic richness and amplify biases} 
While current models' ability to generalize over large amounts of data is one of their biggest assets, their statistical nature and the pressures that shape these models have drawbacks regarding diversity and creativity. Until recently, this might not have been a priority for our field, given that these technologies were largely regarded as domain-specific tools that were often still supervised or post-edited by humans (e.g. chatbots, MT, etc.). However, the fairly recent public release of large, general-purpose, language models raises concerns about the potential implications for language (and language technologies) in the longer term.

From the literature, it seems that indeed, (L)LMs exacerbate imbalances in the data by (i) forgetting low-probability events/words, and (ii) overgenerating high(er) probability ones~\cite{vanmassenhove2021machine,shumailov2023curse}. This could lead to self-reinforcing loops where less frequent linguistic forms and expressions are underrepresented and risk being entirely lost in translation. Low(er)-probability events (words, subwords, etc.) contribute to the complexity and richness across and within language(s), and are important to ensure that models do not converge toward oversimplified, biased language. After all, languages are full of improbable events.

Furthermore, the effects of model collapse are likely not confined to obvious levels such as vocabulary. They may also appear at the morphosyntactic level (e.g., \citet{guo2024benchmarking}), at non-linguistically motivated subword or character levels, in the fact that (the) dominant or target language(s) structure(s) may “bleed through” (e.g., preferences for Subject-Verb-Object constructions), or in the propagation of specific ideologies (e.g., representations of gender). Prior work (e.g. \citet{cao2023multilingual} or \citet{dokic2025mirroring}) has demonstrated how stereotypes encoded in English can “leak” into other languages in multilingual LLMs, with typologically distant languages being particularly vulnerable. This asymmetry is potentially overlooked given the dominance of English in evaluation benchmarks, which could lead to an underestimation of collapse effects in other languages. Similarly, one could hypothesize that multilingual models might start confabulating words similar to English\footnote{\citet{castilho2025synthetic} showed how, just like humans, LLMs tend to resort to what they call "Lazy Gaelicisations" which involve the adaptation of English words towards the Irish ortography. This is, however, also a common strategy among Irish speakers.} or translate English idiomatic expressions literally, even when translating into typologically distant languages, a well-known cause for errors~\cite{karakanta2025close}. 

And last but not least, given the unbalanced nature of language representation on the web, we can furthermore assume that minority, long-tailed and morphologically richer languages are likely affected disproportionaly. As model outputs feed back into future models, and the web increasingly becomes a space where AI systems and agents generate and exchange content with one another, the result is \textbf{a compounding distortion of language use, progressively shifting further away from diversity observed in the real world on the language-internal and cross-linguistic level}.
% \textcolor{teal}{While several mitigation strategies have been proposed to counteract these dynamics, including diversity-aware training objectives and the upweighting of low-frequency phenomena \citep{hashimoto2018fairness,ensign2018runaway,taori2023}, as well as data curation approaches aimed at preserving human-generated content in training pipelines \citep{your_resampling_paper}, such interventions operate within the same underlying optimization and data constraints and therefore may alleviate, but not eliminate, the tendency toward convergence. This is particularly relevant given evidence that alignment and prompting practices can themselves constrain output diversity \citep{yun2025price}.}

\paragraph{Methodological Blind Spots in Measuring Linguistic Diversity}

Current evaluation practices in NLP often involve pairwise comparisons between a single AI-generated text and a single human-written text. Because the model has been trained on massive datasets containing the writing styles, vocabularies, and linguistic patterns of millions of humans, its outputs often exhibit surface-level lexical or syntactic variety that surpasses that of an individual human writer or text~\citep{reviriego2024playing}. This could lead to potentially misleading conclusions, where one might start claiming that AI outputs are \textit{overall} more diverse or more lexically rich than human texts. This way of comparing AI-written vs human-written text \textit{overlooks people's individual variation}.

LLMs can adopt and mimic many different styles, but at inference time, they tend to converge on high-probability patterns. When many generations of outputs are compared over time, these outputs are likely increasingly uniform - a tendency that has already been observed empirically in the form of both intra- and inter-model homogeneity on open-ended questions~\citep{jiang2025artificial}. This can be contrasted against what we observe in human language, which is inherently diverse \textit{across} speakers, contexts, and time. If we instead evaluate diversity at scale (i.e., across many texts or over generations), human-generated language maintains a rich variation through individual idiolects, sociolects, and cultural registers shaped by our individual biases rather than a common one. \textbf{Short-term superficial lexical diversity does not guarantee long-term linguistic sustainability}. Methodological approaches that treat diversity as an isolated textual property risk drawing premature conclusions with respect to diversity in the longer term.

Aside from this methodological blind spot regarding linguistic diversity, it is worth highlighting once more that current evaluation metrics for translation (BLEU~\cite{papineni2002bleu}, METEOR~\cite{banerjee2005meteor}, TER~\cite{snover2006study}, COMET~\cite{rei-etal-2020-comet}) are not designed to capture loss of diversity. Nor should they: diversity is not a property of a single translation, nor even necessarily of a single text. This does not mean, however, that we should ignore it: diversity emerges at the level of systems and over time. Understanding how it evolves across models and generations is important for assessing the broader impact of language technologies. In line with \citet{mccoy2023embers}, I conclude that we should not evaluate LLMs as if they are humans but should instead treat them as a distinct type of system, one that has been shaped by its own particular set of pressures. It is thus important for metrics to be designed in order to reveal \textit{their} idiosyncratic weaknesses.

\paragraph{Compositionality and Systematicity still largely elude LLM capabilities.}

Humans learn language under limited memory capacities; there is a critical period where we can easily learn languages, and our exposure to language is (in some ways) much more restricted than that of LLMs. Our bottlenecks, however, seem to serve as pressure mechanisms for the emergence of structure and compositionality, which allows us, among other things, to create new words that can often immediately be understood by others speaking the same language. We generalize and converge, but we do so towards a productive system. Even though neural networks can behave compositionally and systematically, it is not straightforward for them.

Regarding compositionality in NMT using Transformers~\cite{vaswani2017attention}, recent work by \citet{yin2024compositional} compares the performance of different Transformer-based models (Transformer trained from scratch, pre-trained decoder-only models (BLOOMZ-7b~\cite{muennighoff2022crosslingual} and LLaMA2-13b~\cite{touvron2023llama}) and a pretrained encoder-decoder model (mT5-large~\cite{xue2021mt5}). They show that all these models still struggle when translating new or long compounds. Additionally, lower perplexity source sentences are more likely to be correctly translated into the target, and error rates go up when the length of the compounds increase. These findings are in line with some of the phenomena discussed in the previous section. While they do find that fine-tuned pretrained LLMs outperform Transformer models trained from scratch, they point out that this advantage could be due to pretraining exposure rather than true compositional generalizations. 

In experiments similar to those conducted by \citet{kirby2014iterated} illustrating the effect of cultural transmission through an iterated learning framework, \citet{kouwenhoven-etal-2025-searching} and \citet{kouwenhoven2025shaping} empirically evaluated and compared human-human, LLM-LLM and human-LLM (artificial) language learning to compare how artificial languages differ when optimized by LLMs or humans' inductive biases.\footnote{They do not focus on behavioral biases but on the implicit inductive ones. For humans, these are biases such as preferences for compressibility, simplicity or efficiency), while for LLMs they focused on \textit{increasingly apparent} biases of the Transformer architecture (e.g. simplicity, structure, recency)\cite{kouwenhoven2025shaping}.} Their comparisons of language learning across the three different conditions revealed that, while similar to human vocabularies, LLM languages are subtly different. The LLM optimized languages showed less diversity and variation, making them more \textit{degenerate} in comparison to those optimized for humans. These
differences were alleviated when humans and LLMs collaborated, which underscores that to achieve successful interactions between humans and machines, it is essential to optimize for communicative success since the need to be expressive in human language can prevent convergence.

More generally, \citet{zhou2023data} looked at the link between the complexity of the dataset and the ability of models to generalize. More complex datasets provide: (i) more diversity in terms of the examples the model is exposed to but also, (ii) a reduced repetition preventing the model from \textit{non-generalizable} surface memorization. Yet, as pointed out by \citet{dziri2023_faith_and_fate} but also by \citet{mccoy2023embers}, these models still fail on sometimes surprisingly trivial problems and quickly decay once task complexity increases, indicating once more that these are symptoms of a more fundamental limitation.

\paragraph{Averaging Biases}
Humans are fundamentally biased. For decision-making, we rely on frugal heuristics~\cite{gigerenzer1996reasoning} which are efficient but obviously imperfect. Again, this relates to our cognitive constraints (limited attention, processing capacity, memory). In this context, it is important to highlight that our biases are \textbf{not monolithic}. While it is true that they are partially shaped by our society, environment, and direct surroundings, we each develop a slightly different set of heuristics and biases over time. These are based on our unique experiences and contexts. Regardless of whether they are good or bad, \textit{they are many}. Besides, some of us actively fight or question our own biases when we recognize that they could be harmful, unfair, or simply undesirable.

The biases multilingual LLMs propagate, in contrast, are averaged, dominant ones that are present in an already biased sample of training data. Rather than capturing the diversity and heterogeneity of biases in human reasoning, by letting LLMs drive language change, we risk exacerbating and normalizing dominant biases without having a critical self-reflection or self-correction component.

\paragraph{Invisible Gaps: The Missing Data}
\begin{quote}
    \textit{That which we ignore reveals more than what we give our attention to.}
    
    \hfill --- Mimi Onuoha\footnote{\url{https://github.com/MimiOnuoha/missing-datasets}}
\end{quote}
\vspace{-5pt}
Finally, model collapse is as much about what is not generated as what is. Over-reliance on high-frequency or majority-culture content leads to blind spots in representation. The absence of specific linguistic structures, sociolinguistic variants, or cultural references in training data can render certain forms of expression invisible in model outputs. As models are increasingly retrained on AI-generated content, these omissions risk becoming permanent.

\section{Conclusions}

A few million years ago, we lost our tails. Once functional, they became obsolete and cumbersome. I set out in this paper with the question: Do the many statistical language tails face the same fate, and if so, would we merely be losing the obsolete, fun and decorative elements?

Based on recent LLM-related research, I argue that the long statistical tails of language may indeed face a similar fate. The artificial selection driven by LLMs marks a rather disruptive shift from language being shaped by generational, cultural transmission. Unlike the evolution of language driven by humans, which despite (or because of) our cognitive constraints shaped language into a structured, productive and compositional tool with a rich vocabulary; language shaped by models tends to collapse towards what is likely, driven by statistical biases. The interplay and balance between convergence and divergence, that characterizes human behavior and communication on multiple levels, risks being lost. Unlike humans, models are not intrinsically motivated to be creative, to express belonging or differences, or to innovate. There is no capacity for critical self-reflection or self-correction, and only a limited plurality of voices. 

As these systems increasingly \textit{ingest} their own or other models' outputs, the risk of flattening linguistic diversity grows, with rare words, less-resourced languages, and culturally significant variation most at risk. Preserving the long tails of language means rethinking how we evaluate and train these systems, not just for accuracy or fluency, but for the communicative richness that makes language human. \textbf{Without our tails, we risk losing the balancing act} by converging, collapsing, and flattening the expressive multilingual linguistic, social and cultural diversity.

\section*{Limitations}
The arguments presented in this paper are intended to provoke critical reflection on the trajectory of language in the era of LLMs; however, they are subject to several limitations regarding scope, empirical generalization, and my own perspective. While I draw on a synthesis of recent findings in model collapse (and more specifically `diversity collapse'), iterated learning, and sociolinguistics, the long-term impact of LLMs on natural human language and speech remains a developing phenomenon. The limitations and concerns discussed reflect the current state-of-the-art. It is possible that future architectures, perhaps those incorporating neuro-symbolic reasoning or novel inductive biases, can mitigate certain aspects of model collapse. My position is therefore a critique of the current trajectory of generative AI rather than an immutable law of artificial/machine intelligence.

I furthermore acknowledge an inherent selection bias in the literature reviewed and the examples provided. My perspective is situated within an academic tradition that values linguistic richness. I recognize that researchers from more functionalist or engineering-driven backgrounds might interpret the "flattening" of language as an increase in communicative efficiency or standardization rather than a loss of richness.

% \section*{Acknowledgments}
% I thank the anonymous ARR reviewers for their thoughtful, encouraging, and constructive feedback. In line with the ACL policy on AI writing assistance, AI tools were used solely to improve the English of this paper (e.g., correcting grammatical errors and suggesting alternative phrasings). No AI tools were used to generate scientific content.

\section*{Acknowledgments}
I thank the anonymous ARR reviewers for their thoughtful, encouraging, and constructive feedback. I am grateful to Afra Alishahi for feedback on an earlier draft, and to Dimitar Shterionov for his feedback, guidance, and many years of discussions that have shaped my thinking on this topic. I also thank my father for introducing me to Simon Kirby’s work via a BBC documentary years ago, and for the steady stream of book recommendations ever since. Finally, I thank the MT Summit 2025 organizers for the opportunity to present an earlier version of these ideas as a keynote, and the audience for their insightful comments and discussions. In line with the ACL policy on AI writing assistance, AI tools were used solely to improve the English of this paper (e.g., correcting grammatical errors and suggesting alternative phrasings). No AI tools were used to generate scientific content.

\bibliography{anthology,custom}
\bibliographystyle{acl_natbib}

\end{document}